\documentclass[runningheads]{llncs}
\usepackage[T1]{fontenc}
\usepackage{graphicx}
\usepackage{booktabs}
\usepackage[misc]{ifsym}
\newcommand{\corr}{(\Letter)}
\usepackage{mwe}
\usepackage{microtype}
\usepackage{graphicx}

\usepackage{booktabs} 

\usepackage{cite}
\usepackage{float} 
\usepackage{subcaption} 
\usepackage{multirow} 
\usepackage{rotating} 
\usepackage{tabularx} 
\usepackage{graphicx} 

\usepackage{caption}
\usepackage{mathtools}
\usepackage{amsfonts}
\usepackage{dsfont}
\usepackage{algorithm}
\usepackage{algorithmic, eucal}

\usepackage{url}
\usepackage{xcolor}

\usepackage{bbding}

\definecolor{cvprblue}{rgb}{0.21,0.49,0.74}
\usepackage[breaklinks,colorlinks,citecolor=cvprblue]{hyperref}

\usepackage{color}

\makeatletter
\newcommand{\printfnsymbol}[1]{%
  \textsuperscript{\@fnsymbol{#1}}%
}
\makeatother

\begin{document}

\title{Self-Supervised Spatial-Temporal Normality Learning for Time Series Anomaly Detection}

\titlerunning{Spatial-Temporal Normality Learning for Anomaly Detection}

\author{Yutong Chen\inst{1,2}\thanks{Equal contribution} \and
Hongzuo Xu\inst{3}\printfnsymbol{1} \and Guansong Pang\inst{4} \corr \and Hezhe Qiao\inst{4} \and \\ Yuan Zhou\inst{3} \and Mingsheng Shang\inst{1,2} \corr}

\authorrunning{Y. Chen et al.}

\institute{Chongqing Institute of Green and Intelligent Technology, \\ Chinese Academy of Sciences, Chongqing 400714, China \\
\and
Chongqing School, University of Chinese Academy of Sciences, \\ Chongqing 400714, China \\
\email{\{chenyutong,msshang\}@cigit.ac.cn}
\and
Intelligent Game and Decision Lab (IGDL), Beijing 100091, China \\ 
\email{hongzuoxu@126.com, yuaanzhou@outlook.com}
\and
Singapore Management University, 178902, Singapore \\
\email{gspang@smu.edu.sg, hezheqiao.2022@phdcs.smu.edu.sg}
}

\maketitle             

\begin{abstract}
Time Series Anomaly Detection (TSAD) finds widespread applications across various domains such as financial markets, industrial production, and healthcare. Its primary objective is to learn the normal patterns of time series data, thereby identifying deviations in test samples. 
Most existing TSAD methods focus on modeling data from the temporal dimension, while ignoring the semantic information in the spatial dimension.
To address this issue, we introduce a novel approach, called \underline{S}patial-\underline{Te}mporal \underline{N}ormality learning (\textbf{STEN}). STEN is composed of a sequence Order prediction-based Temporal Normality learning (OTN) module that captures the temporal correlations within sequences, and a Distance prediction-based Spatial Normality learning (DSN) module that learns the relative spatial relations between sequences in a feature space. By synthesizing these two modules, STEN learns expressive spatial-temporal representations for the normal patterns hidden in the time series data. 
Extensive experiments on five popular TSAD benchmarks show that STEN substantially outperforms state-of-the-art competing methods.
Our code is available at \renewcommand\UrlFont{\color{blue}}\url{https://github.com/mala-lab/STEN}.

\end{abstract}

\section{Introduction}
Time series data are pervasive in many real-world application domains, including finance, industrial production, network traffic, and health monitoring \cite{Anandakrishnan_Kumar_Statnikov_Faruquie_Di_2017,Ren_Xu_Wang_Yi_Huang_Kou_Xing_Yang_Tong_Zhang_2019,pereira2019learning}. 
Within these time series data, there can exist exceptional data observations, a.k.a., anomalies, that deviate significantly from the majority of data, which often indicates an abnormal change of the data-generating mechanism \cite{darban2022deep}. 
These anomalies are of great interest to the analysts since accurately detecting them is significant in alarming faults in target systems and preventing potential losses.

Anomaly detection aims to learn data normality during training and identify those exceptional data during inference \cite{pang2021deep}. It is a non-trivial task to precisely learn the normality of time series data. The challenge is primarily due to the unsupervised nature of anomaly detection. 
Models like neural networks are actuated by supervisory signals to learn high-level patterns from given training data. However, it is prohibitively costly to collect large-scale labeled data for anomaly detection. Thus, learning from unlabeled data is generally more preferable yet it is difficult due to the lack of labeled training data. 
The challenge is aggravated by the inherent complexity of time series data. Time series data is characterized by multiple components from temporal and spatial perspectives, including the trending and seasonal changes along the time dimension and the spatial proximity relation of temporal sequences in the feature space. 

Current time series anomaly detection (TSAD) methods typically leverage an encoder-decoder architecture and assess data abnormality according to reconstruction/prediction errors of testing data \cite{Xu_Wu_Wang_Long_2021,xu2022calibrated,yang2023dcdetector,Wu_He_Lin_Su_Cui_Maple_Jarvis_2022,Ren_Xu_Wang_Yi_Huang_Kou_Xing_Yang_Tong_Zhang_2019,deng2021graph,chen2024lara}. 
Despite their general effectiveness on various datasets, these methods often tend to overfit the training data and fail to distinguish anomalies from the normal sequences, since both types of sequences have similarly small reconstruction/prediction errors. 
Self-supervised learning 
has been emerging as a promising technique in anomaly detection due to its capability of deriving supervisory signals from the data itself. Existing self-supervised methods, e.g., via pretext tasks like association learning \cite{Xu_Wu_Wang_Long_2021} and dual attention contrastive representation learning \cite{yang2023dcdetector}, successfully learn discriminative models to capture normal semantics of the temporal continuity into the representation space. 
However, 
apart from the data regularities reflected in the temporal dimension, spatial normality is also a desideratum of comprehensive normality learning in time series data. 

\begin{figure}[htbp]
\includegraphics[width=\textwidth]{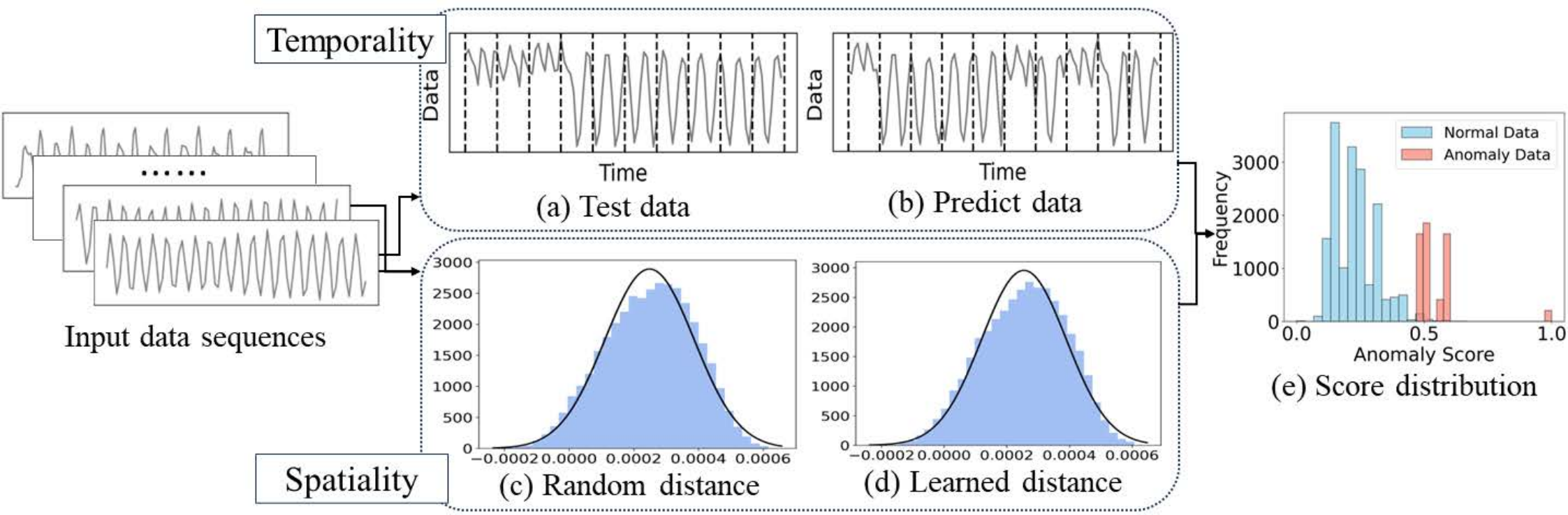}
\caption{Our key insights. (a) A sequence containing abnormal data is divided into equal-length sub-sequences by dashed lines, with anomalies presented in the first three sub-sequences. (b) Sub-sequences arranged according to the predicted order distribution, showing differences between the predicted order distribution and the original data due to the presence of the anomaly. (c) The distribution of distances between sequence pairs in a random projection space, which can well preserve the spatial information of the sequences within the feature space. (d) The distribution of distances between sequence pairs learned by a trainable network, effectively resembling the distance distribution. (e) The distribution of anomaly scores obtained by considering the prediction discrepancies in both temporal and spatial dimensions for normal and abnormal data. 
The results are based on the Epilepsy dataset~\cite{xu2022calibrated}.} 
\label{fig:Insight}
\end{figure}

In light of this limitation, this paper investigates an intriguing question: \textit{Can we simultaneously learn 
spatial-temporal normality of time series data in the self-supervised paradigm?}
As shown in Fig. \ref{fig:Insight}, the input data sequences are the training data of the Epilepsy dataset~\cite{xu2022calibrated}, which illustrates the normal patterns.
Time series continuously change along the time dimension, and this continuity can be fully leveraged in temporal normality learning. More specifically, after splitting the raw time sequences into several sub-sequences, the normality can be modeled by learning the sub-sequence order.
During inference, giving a testing sequence with anomalies (Fig. \ref{fig:Insight}(a)), the learning model ranks the randomly shuffled sub-sequences and finally yields a sequence that resembles the training data (Fig. \ref{fig:Insight}(b)). This predicted order depicts a large difference compared to the original distribution for the anomalous sequence, whereas the difference is small for normal sequences.   

In terms of spatial normality, we investigate the pairwise distance of time series data sequences in a projection space. Generally, it is challenging to obtain the spatial relation of data samples evolving along the time, especially under the unsupervised learning paradigm. Motivated by the effective spatial information preservation by random projection~\cite{bingham2001random,hegde2007random,wang2019unsupervised}, we leverage the pairwise distance in a randomly projected feature space. Since normal sequences often locate in a dense region, 
as shown in Fig. \ref{fig:Insight}(c), the distance distribution of a normal sequence typically follows a Gaussian-like distribution with the expected value approximating zero. On the contrary, that of anomalous sequences tends to have a markedly different distance-based spatial distribution (e.g., uniform-like distribution). 
Neural networks are capable of predicting those distance-based spatial information, 
as shown in Fig. \ref{fig:Insight}(d). By learning to accurately predict these distances, the networks learn the spatial distribution of the sequence normality. 
During inference, testing anomalous sequences manifest significant deviation from the normal data distribution according to the anomaly score (see Fig. \ref{fig:Insight}(e)). 

Based on this insight, we propose a novel approach for TSAD, called \underline{S}patial-\underline{Te}mporal \underline{N}ormality learning (\textbf{STEN} for short), in which two pretext tasks are designed to construct a self-supervised learning pipeline for the joint spatial-temporal normality modeling. 
Specifically, we first devise an Order prediction-based Temporal Normality learning module (OTN). The inputting time sequences are split into several sub-sequences. Supervised by their genuine order, the neural network learns how to sort shuffled sub-sequences, during which the temporal continuity can be captured and represented. On the other hand, we further introduce a Distance prediction-based Spatial Normality learning module (DSN), which models the spatial proximity of sequences in the representation space. 

The main contributions are summarized as follows. 
\begin{itemize}
    \item We introduce the concept of spatial-temporal normality of time series data and propose a novel anomaly detection approach STEN. STEN achieves spatial-temporal modeling of the data normality, contrasting to the existing methods that are focused on the temporal perspective only.
    \item To learn the temporal normality, we propose the OTN module. 
    This module captures temporal associations and contextual information in time series by learning the distribution of sub-sequence orders.
    Different from many Transformer-based methods that calculate associations via heavy attention mechanisms, our method successfully models temporal normality by fully harnessing the unique continuity of time series data. 
    \item To learn the spatial normality, we propose the DSN module. By wielding an informative random projection space, our approach further restrains spatial proximity in the representation space. Compared to mainstream reconstruction-based models that focus on individual normality, our approach investigates data affinity among the sequences, thereby capturing the normality beyond the temporal perspective.
    \item STEN outperforms state-of-the-art methods on five popular benchmark datasets for time series anomaly detection. 
\end{itemize}

\section{Related Work}

\noindent\textbf{Time Series Anomaly Detection.}
Time series anomaly detection is an old discipline, which has received increasing attention in recent years. 
Early traditional methods focus on statistical approaches such as moving averages, AutoRegressive Integrated Moving Average (ARIMA) models~\cite{Box_Pierce_1970}, and their multiple variants~\cite{Li_Di_Shen_Chen_2021}. With the emergence of machine learning technology, techniques including classification~\cite{Karczmarek_Kiersztyn_Pedrycz_Al_2020}, clustering~\cite{hautamaki2004outlier}, ensemble learning~\cite{xu2023dif},and time series forecasting~\cite{gunnemann2014robust} are applied to time series anomaly detection. 
Besides, Tsfresh has inspired the window-to-feature approach, enhancing the efficiency of feature extraction in time series analysis~\cite{christ2018time}. ROCKET's focus on sub-sequence patterns through random convolutional kernels has inspired advancements in capturing local temporal patterns~\cite{dempster2020rocket}. 
However, traditional methods are often constrained by the learning capability and the quantity of labeled data, making it challenging to achieve satisfactory performance.

With the burgeoning of deep learning techniques, many deep anomaly detection methods have been introduced in the literature. 
Owing to deep learning's powerful capability in modeling intricate data patterns and distributions, deep anomaly detection methods dramatically improve the detection performance over conventional methods~\cite{pang2021deep}.  
Mainstream deep time-series anomaly detection methods are based on generative models, in which the learning models are trained to predict or reconstruct original raw time series data~\cite{darban2022deep}. Prediction-based methods train models to forecast the value of the next timestamp, using the discrepancy between predicted and actual values to indicate the abnormal degree of the current timestamp. 
Reconstruction-based methods compare the error between reconstructed and actual values. 
These methods employ various neural network architectures such as internal memory, dilated convolutions, and graph structure learning, intending to capture the temporal characteristic of time series, yielding impressive results across numerous benchmarks~\cite{Wu_He_Lin_Su_Cui_Maple_Jarvis_2022,Ren_Xu_Wang_Yi_Huang_Kou_Xing_Yang_Tong_Zhang_2019,deng2021graph,chen2024lara}.
Additionally, some studies have coupled them with adversarial training to amplify the discriminability of anomalies~\cite{Audibert_Michiardi_Guyard_Marti_Zuluaga_2020,Liu_Zhou_Ding_Hooi_Zhang_Shen_Cheng_2023}. 
A recent work, named Anomaly Transformer~\cite{Xu_Wu_Wang_Long_2021}, utilizes the Transformer to model the associations between sequences and their neighboring priors, identifying anomalies through association differences.

\vspace{0.2cm}
\noindent\textbf{Self-supervised Learning on Time Series.} Self-supervised learning generates supervision signals from the data itself. Via various proxy learning tasks, the learning models embed data patterns into the projection space, offering semantic-rich representations to downstream tasks. 
Self-supervised time series analysis can be categorized into generative-, contrastive-, and adversarial-based methods~\cite{zhang2023self}.
As introduced above, generative methods rely on prediction/reconstruction learning objectives to capture the characteristics of time series data. These methods often overfit the training data, and anomalies may have similarly small reconstruction errors. 
Adversarial-based techniques, noted in~\cite{Zhou_Liu_Hooi_Cheng_Ye_2019,li2022tts}, might be hampered by complex and unstable training regimes.

Contrastive learning defines positive and negative pairs, and the learning process minimizes the distance between positive pairs and maximizes the distance between negative pairs in the feature space. 
This approach can encourage the model to learn meaningful and discriminative features. 
This concept has been applied in studies such as~\cite{schneider2022detecting,Yue_Wang_Duan_Yang_Huang_Tong_Xu_2022,sun2023unraveling}. 
DCdetector~\cite{yang2023dcdetector} is one of the latest frameworks devised for anomaly detection tasks, which employs a dual-attention contrastive representation mechanism to differentiate between normal and abnormal samples effectively. 
However, this approach still overlooks the importance of spatial normality learning in comprehending the normal patterns of time series data.

\section{STEN: Spatial-Temporal Normality Learning}
\subsection{Problem Statement}
Let $\mathcal{X}=\left\langle\mathbf{x}_1, \mathbf{x}_2, \cdots, \mathbf{x}_N\right\rangle$ be a sequence of multivariate time series with $N$ observations, with each observation $\mathbf{x}_{t} \in \mathbb{R}^{D}$ denotes the data values of $D$ variants at a certain timestamp $t$, where $1 \leq t \leq N$, then unsupervised time series anomaly detection (TSAD) aims to learn an anomaly scoring function $f$ without reliance on any labels of the training data $\mathcal{X}$. $f$ is applied to measure the abnormality of observations in testing sequences $\mathcal{X}_{\text{test}}$. 
Higher anomaly scores indicate a higher likelihood to be anomalies. In STEN, we design a particular $f$ that can effectively capture spatial-temporal normality in training data $\mathcal{X}$.

\subsection{Overview of The Proposed Approach}

\begin{figure}[t]
\includegraphics[width=\textwidth]{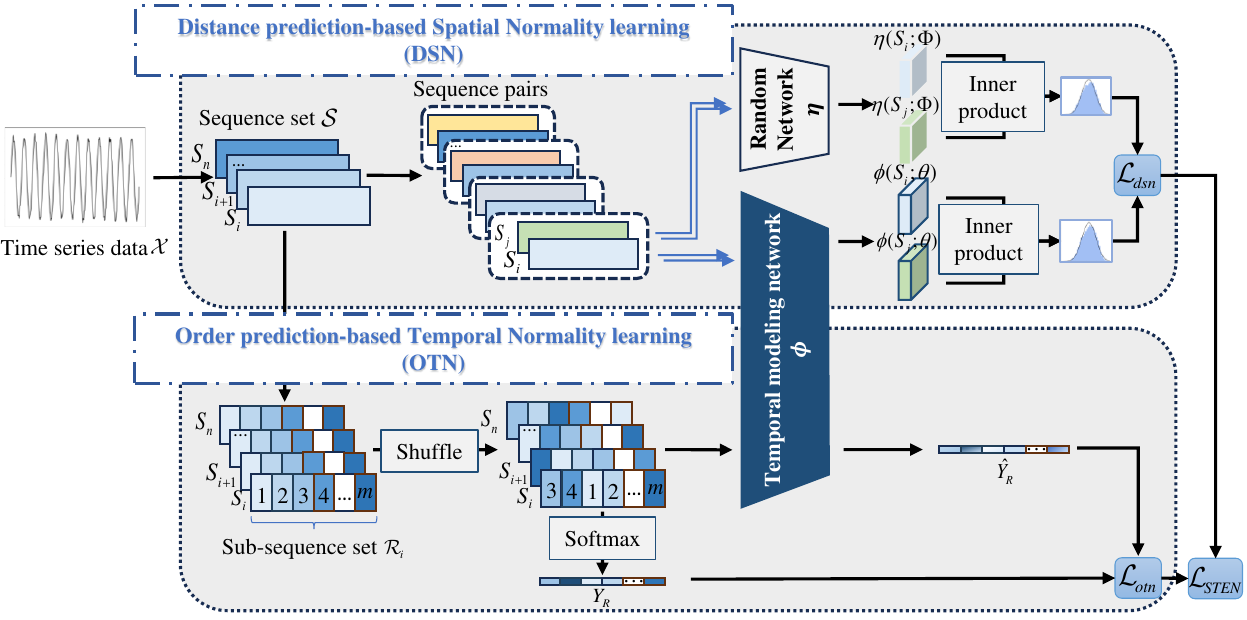}
\caption{Overview of STEN. STEN consists of two self-supervised components: 
DSN and OTN.
In DSN, the distances between sequence pairs after being projected by a random network $\eta$ form a compact distribution, from which we distill the spatial normality patterns using a
MSE loss function $\mathcal{L}_{dsn}$. 
OTN captures temporal normality by predicting the order among sub-sequences after a random shuffling using a distribution similarity-based loss function, Jensen-Shannon divergence.}\label{fig: framework}
\end{figure}

This paper proposes a novel unsupervised TSAD method STEN, in which data normality is learned from both temporal and spatial perspectives. 
As shown in Fig. \ref{fig: framework}, STEN consists of an Order prediction-based Temporal Normality learning module (OTN) and a Distance prediction-based Spatial Normality learning module (DSN). 
OTN learns the temporal normality information of the time series by modeling the sequential distribution of sub-sequences. On the other hand, DSN is designed to utilize the distance prediction between the sequence pairs for spatial normality learning. 
Finally, the anomaly score which reflects both temporal and spatial normality can be used to distinguish normal and abnormal sequences. 
The details of OTN and DSN are discussed in the following two sections.

\subsection{OTN: Order Prediction-based Temporal Normality Learning}

Time series datasets are essentially organized into specific sequences, where connectivity and sequential information of sub-sequences illuminate time-related patterns and contextual relationships within the data. The temporality of time series data reflects the dynamic changes of data in the time dimension. Analyzing temporality can help understand the dynamic changes and time correlation of data.  Therefore, the OTN module is designed to extract temporal normality in time series data by predicting the temporal order between their sub-sequences. 

To this end, OTN models the temporal patterns by predicting the primary temporal order of randomly shuffled sub-sequences. Specifically, a sliding window of length $L$ and stride $R$ are first used to divide $\mathcal{X}$ into a collection of $n$ sequences ${\cal S} =\left \{ S_{1},S_{1+R},...,S_{n}  \right \}$ , where each $S_{t}=\left \langle  \mathbf{x}_{t},\mathbf{x}_{t+1},...,\mathbf{x}_{t+L-1}    \right \rangle $.
For each sequence $S \in {\cal S_i}$,  we generate a fixed number of equal-length short sequences with a fixed length $l$ and stride $r$. One sequence $S_{t}$ will result in $m$ sub-sequences ${\cal R}_{t} =\left \{ R_{1},R_{2},...,R_{m}    \right \} $. The position orders of the sub-sequences are used as a set of self-supervised class labels to train the neural network $\phi $. 
The Jensen-Shannon (JS) divergence, a popular method for measuring the difference between two different distributions, is then used to define our self-supervised loss $\mathcal{L}_{otn}$ in predicting the order distributions of these sub-sequences, which is formulated as follows: 

\begin{equation}
{{\cal L}_{{\rm{otn }}}} = \sum\limits_{i = 1}^c  {{{\hat Y}_{{R_i}}}} \log \left( {\frac{{ {{{\hat Y}_{{R_i}}}} }}{{\frac{1}{2}\left( { {{{\hat Y}_{{R_i}}}}  + {Y_{R_i}}} \right)}}} \right) + \sum\limits_{i = 1}^c {{Y_{R_i}}} \log \left( {\frac{{{Y_{{R_i}}}}}{{\frac{1}{2}\left( { {{{\hat Y}_{{R_i}}}}  + {Y_{{R_i}}}} \right)}}} \right),
\end{equation}
where $\hat{Y}_{R}$ represents the softmax results of the temporal order prediction for the sub-sequence set $\mathcal{R}$ and $Y_{{R}}$ is the ground truth order distribution. Minimizing this loss enables the learning of temporal normal patterns based on the local sub-sequence context.

\subsection{DSN: Distance Prediction-based Spatial Normality Learning}

The spatial patterns of time series data we aim to capture are the spatial distribution of the data that changes over time in a feature space. It describes the spatial distribution patterns, trends, and changes of the data over a period of time, which cannot be captured by the temporal modeling. 
To complement the temporal normality learning, we propose the DSN module to model the spatial normality of the time series data in a feature space. Different from OTN that is designed to learn the temporal dynamics within the individual sequences, DSN is designed to model the spatial relation between the sequences, and thus, it is performed on the sequences set $\cal S$ rather than the sub-sequences.  

One challenge here is the lack of supervision signals for learning the spatial patterns. Inspired by the solid theoretical and empirical results in \cite{bingham2001random,hegde2007random,wang2019unsupervised}, we propose to use the distance of the sequences in a feature space spanned by random projection as our supervision source for learning the spatial patterns among the sequences. It has been theoretically justified and empirically shown in these prior studies that although it is in a randomly projected feature space, the distance information in the original space can be well preserved. To utilize these random distances through deep neural networks, we design the self-supervised distance prediction method. 
More specifically, for each sequence $S_i$, we randomly select one sequence from the remaining sequences and construct a sequence pair $({S}_i, {S}_j)$. 
They are then respectively fed to a trainable network $\phi :\mathbb{R} ^{D} \mapsto \mathbb{R} ^{M}$ with parameters $\Theta$ and a random network $\eta :\mathbb{R} ^{D} \mapsto \mathbb{R} ^{M}$, which is a representation learner with parameters $\Phi$. The parameters $\Phi$ in the network $\eta$ is initialized with random weights and frozen, which serves as the random project model. It is used to project the original sequences and obtain the distance-based relation via:

\begin{equation}
   {d_{\eta }}(S_i, S_j) =\eta (S_{i};\Phi  )^{T} \eta (S_{j}; \Phi  ),
\end{equation}
where ${d_{\eta }}(S_i, S_j)$ represents a spatial relation of the two sequences in the projected feature space. To learn these spatial relations, we train the network $\phi$ to predict/distill these distances via:

\begin{equation}
   \mathcal{L}_{dsn} =  \frac{1}{n} \sum_{i=1}^{n} ({d_{\phi }}(S_i, S_j)- {d_{\eta  }}(S_i, S_j) )^{2},
\end{equation}
where $n$ is the number of sequence pairs and ${d_{\phi }}(S_i, S_j)$ is defined as follows: 

\begin{equation}
   {d_{\phi }}(S_i, S_j)=\phi (S_{i};\Theta  )^{T} \phi (S_{j};\Theta  ).
\end{equation}

The training of the network $\phi$ can be seen as distilling semantic information from the distance of sequence pairs, learning the relative spatial normality of time series data. A normal sequence typically include a dense set of normal sequences in their local neighborhood while being distant from the other normal sequences. Thus, its distance-based spatial relation to the other sequences forms a Gaussian-like distribution. By contrast, the anomalous sequences are assumed to be distant from most sequences, so its distribution of the distances to other sequences tends to be a uniform one. DSN is designed to learn such discriminative patterns by minimizing the loss $\mathcal{L}_{dsn}$.

\subsection{Training and Inference}

\noindent \textbf{Training.} 
The OTN and DSN are synthesized in our approach STEN to capture the spatial-temporal dependencies of the time series data, offering a comprehensive modeling of the normal patterns. Thus, the overall loss function in STEN is composed of the loss functions from the two above two self-supervised tasks: 
\begin{equation}
   \mathcal{L}_{STEN}= \mathcal{L}_{otn}+\alpha  \mathcal{L}_{dsn},
\end{equation}
where $\alpha$ is a hyperparameter used to modulate the two modules, and the learnable parameters in network $\phi $ are jointly learned by the loss function $\mathcal{L}_{STEN}$. 

\noindent \textbf{Inference.}
In the OTN module, because of the continuity of the normal patterns, the order distribution reflects the overall order prediction of a sub-sequence collection $\cal{R}_{t}$, and the discrete order of a single sub-sequence $R_{i}$ reflects the prediction of the current sub-sequence. 
Thus, the discrepancies in both types of predictions for normal data will be less than those for anomalies. Motivated by this, we use both of these differences in our anomaly scoring, which is defined as follows:
\begin{equation}\label{eqn:otn_score}
{{Score}_{otn}}(R_i) = \frac{\left| { {{{\hat Y}_{{R_{i}}}}}  - {Y_{{R_i}}}} \right|}{{\mathcal{L}}_{{otn}_{i}}},
\end{equation}
where the numerator is the difference in a single subspace while the denominator summarizes the differences across all sub-sequences. 
Note that ${\cal L}_i$ has a different scale from $\mathcal{L}_{otn}$, so we perform an upsampling of this score in Eq. (\ref{eqn:otn_score}) by replicating it from a scalar to a vector of length of $m$.

The spatial normality of the sequences is captured in $\mathcal{L}_{dsn}$, i.e., for a normal sequence $S_i$, it is expected to have substantially smaller $\mathcal{L}_{dsn}$ than anomalous sequences when paired with other sequences $S_j$. To utilize spatial-temporal normality for TSAD, STEN defines an overall anomaly score as:
\begin{equation}
    Score(R_i)={{Score}_{otn}}(R_i) +\beta {Score}_{{dsn}}(R_{i}),
\end{equation}
where $\beta$ is a hyperparameter to control the importance of the spatial normality term ${Score}_{{dsn}}(R_{i})=\mathcal{L}_{dsn}(R_{i})$. Note that $\mathcal{L}_{dsn}$ measures the normality at the sequence level, and we assign the same anomaly score for the sub-subsequences within a sequence.

\section{Experiments}

\subsection{Experimental Setup}

\noindent \textbf{Benchmark Datasets.} Five publicly available multivariate time series datasets are used in our experiments, with the relevant statistics shown in Table \ref{tab:dataset}. 
PSM (Pooled Server Metrics Dataset) ~\cite{abdulaal2021practical} is a dataset of IT system monitoring signals from eBay server machines with 25 dimensions.
Both MSL (Mars Science Laboratory Dataset) and SMAP (Soil Moisture Active Passive Dataset)~\cite{Hundman_Constantinou_Laporte_Colwell_Soderstrom_2018} are public datasets collected by NASA with 27 and 55 entities respectively, which contain sensor and actuator data from the Mars Rover, as well as soil samples and telemetry information from a satellite.
Epilepsy (Epilepsy seizure dataset) is an activity dataset collected from a triaxial accelerometer on the wrist of a human subject's hand, and we treat data during walking, running, and sawing as normal data and seizures as anomalies according to~\cite{xu2022calibrated}. 
DSADS (Daily and Sports Activities Dataset) collects motion sensors from eight subjects, including 19 daily and physical activities. Following~\cite{xu2022calibrated}, we use intense activities as the anomaly class, with the data collected from other activities used as normal.

\begin{table}[t]
\centering
\caption{Key dataset statistics.}
\label{tab:dataset}
\scalebox{0.8}{
\begin{tabular}{lccccc}
\toprule
 & $N_{train}$ & $N_{test}$ & \#Dimensions & \#Entities & AnomalyRatio(\%) \\
\midrule
PSM &  132,481 & 87,841 & 25 & 1 & 27.8 \\
MSL &  2,160 & 2,731 & 55 & 27 & 10.7 \\
SMAP & 2,556 & 8,071 & 25 & 55 & 13.1 \\
Epilepsy & 33,784 & 22,866 & 5 & 1 & 10.2 \\
DSADS & 85,500 & 57,000 & 47 & 1 & 6.3 \\
\bottomrule
\end{tabular}}
\end{table}

\vspace{0.3cm}
\noindent \textbf{Competing Methods.} STEN is compared with the following eight state-of-the-art (SotA) anomaly detectors specifically designed for time series data.
These competing methods can be generally categorized into four types, i.e., (1) reconstruction-based models: MSCRED (MSC for short)~\cite{zhang2019deep}, TranAD (Tran for short)~\cite{tuli2022tranad}, and AnomalyTransformer (AT for short)~\cite{Xu_Wu_Wang_Long_2021}; 
(2) forecasting-based models: GDN \cite{deng2021graph}; (3) one-class classification models: TcnED (TED for short)~\cite{Garg_Zhang_Samaran_Savitha_Foo_2022}, COUTA (COU for short)~\cite{xu2022calibrated}, and NCAD (NCA for short)~\cite{carmona2021neural}; (4) contrastive learning-based models: DCdetector (DC for short)~\cite{yang2023dcdetector}.

\begin{table}[htbp]
  \centering
  \caption{AUC-ROC, AUC-PR, and $F_{1}$ results on five TSAD datasets.}
   \scalebox{0.8}{%
    \begin{tabularx}{\textwidth}{cl
    >{\centering\arraybackslash}X
    >{\centering\arraybackslash}X
    >{\centering\arraybackslash}X
    >{\centering\arraybackslash}X
    >{\centering\arraybackslash}X
    >{\centering\arraybackslash}X
    >{\centering\arraybackslash}X
    >{\centering\arraybackslash}X
    >{\centering\arraybackslash}X}
    \toprule
            & \textbf{Dataset} & \textbf{Ours} & \textbf{DC} & \textbf{AT} & \textbf{NCA} & \textbf{Tran} & \textbf{COU} & \textbf{TED} & \textbf{GDN} & \textbf{MSC} \\
    \midrule
    \multirow{5}[2]{*}{\begin{sideways}\textbf{AUC-ROC}\end{sideways}} & PSM   & \textbf{0.998 } & 0.967  & 0.993  & 0.973  & 0.972  & 0.975  & 0.970  & 0.968  & 0.963  \\
          & MSL   & \textbf{0.996 } & 0.961  & 0.979  & 0.983  & 0.986  & 0.989  & 0.983  & 0.976  & 0.899  \\
          & SMAP  & \textbf{0.999 } & 0.991  & 0.993  & 0.970  & 0.921  & 0.919  & 0.926  & 0.978  & 0.821  \\
          & Epilepsy & \textbf{0.998 } & 0.886  & 0.996  & 0.984  & 0.935  & 0.951 & 0.933  & 0.990  & 0.832  \\
          & DSADS & \textbf{0.994 } & 0.983  & 0.962  & 0.985  & 0.984  & 0.992  & 0.985  & 0.974  & 0.905 \\
    \midrule
    \multirow{5}[2]{*}{\begin{sideways}\textbf{AUC-PR}\end{sideways}} & PSM   & \textbf{0.995 } & 0.946  & 0.989  & 0.942  & 0.955  & 0.955  & 0.948  & 0.940  & 0.932  \\
          & MSL   & \textbf{0.941 } & 0.915  & 0.933  & 0.868  & 0.905  & 0.886  & 0.897  & 0.849  & 0.483  \\
          & SMAP  & \textbf{0.991 } & 0.961  & 0.971  & 0.884  & 0.753  & 0.758  & 0.717  & 0.869  & 0.644  \\
          & Epilepsy & \textbf{0.991 } & 0.837  & 0.980  & 0.943  & 0.790  & 0.760  & 0.773  & 0.967  & 0.565  \\
          & DSADS & \textbf{0.948 } & 0.876  & 0.926  & 0.880  & 0.917  & 0.947  & 0.913  & 0.785  &  0.659 \\
    \midrule
    \multirow{5}[2]{*}{\begin{sideways}\textbf{BEST $F_{1}$}\end{sideways}} & PSM   & \textbf{0.986 } & 0.959  & 0.982  & 0.908  & 0.914  & 0.925  & 0.881  & 0.870  & 0.889  \\
          & MSL   & \textbf{0.944 } & 0.932  & 0.942  & 0.809  & 0.888  & 0.867  & 0.890  & 0.852  & 0.605  \\
          & SMAP  & \textbf{0.974 } & 0.963  & 0.967  & 0.845  & 0.699  & 0.701  & 0.711  & 0.781  & 0.668  \\
          & Epilepsy & \textbf{0.991 } & 0.875  & 0.987  & 0.904  & 0.803  & 0.793  & 0.777  & 0.918  & 0.640  \\
          & DSADS & \textbf{0.934 } & 0.908  & 0.932  & 0.875  & 0.846  & 0.901  & 0.844  & 0.825  &  0.657 \\
    \bottomrule
    \end{tabularx}%
    }%
  \label{tab:main_result}%
\end{table}%

\vspace{0.3cm}
\noindent \textbf{Evaluation Metrics.}
The performance of STEN is measured according to a wide range of evaluation metrics. 
Following the mainstream studies in this research line~\cite{Audibert_Michiardi_Guyard_Marti_Zuluaga_2020,tuli2022tranad,xu2022calibrated}, we adopt three popular metrics including the Area Under the Receiver Operating Characteristic Curve (AUC-ROC), the Area Under the Precision-Recall Curve (AUC-PR), and the best $F_{1} $ score. Note that these three metrics are calculated upon anomaly scores processed by the point-adjust strategy~\cite{Xu_Feng_Chen_Wang_Qiao_Chen_Zhao_Li_Bu_Li_et_al._2018}. 
It is a commonly used strategy in time series anomaly detection~\cite{carmona2021neural,xu2022calibrated,yang2023dcdetector,Xu_Wu_Wang_Long_2021}, which adjusts the anomaly score of each anomaly segment to the highest score within this segment. 
To circumvent biases introduced by point adjustment, we further employ five recently proposed evaluation measures.  
Considering the temporal relationship/distance between ground truths and predictions, \cite{huet2022local} calculates the affiliation precision and recall. 
We utilize the harmonic mean of these two measures, i.e., Affiliation $F_{1}$ (denoted by Aff-$F_{1}$). 
\cite{paparrizos2022volume} introduces a novel metric called Range-AUC, which extends the AUC measurement to account for range-based anomalies. Additionally, the paper introduces volume under the ROC surface (VUS-ROC) and volume under the PR surface (VUS-PR) as new metrics for computing the volume under ROC and PR curves, respectively.

\vspace{0.3cm}
\noindent \textbf{Implementation Details.}
We summarize the default implementation settings of our method STEN as follows. 
Both the number of generated sub-sequences $m$ and the sub-sequence length $l$ are set to 10 across all datasets. 
The temporal modeling network $\phi$ is composed of a single-layer GRU network with shared parameters across 10 units, and the dimension of the hidden state $d_{model} $ is set to 256. To balance the influence of OTN and DSN,  the hyperparameters $\alpha $ and $\beta$ are set to $1$ by default.  
For calculating the affiliation metric, an anomaly is defined as any timestamp whose anomaly score exceeds the $(100 - \delta)$ percentile, with $\delta $ setting to 0.6 by default.  The weight parameters are optimized using Adam optimizer with a learning rate of $10^{-5} $ for a total of 5 epochs.
For the PSM, SMAP, and DSADS datasets, the batch size is set to 256, while for the MSL and Epilepsy datasets, each mini-batch contains 64 training samples.

\subsection{Main Results}
To verify the detection performance of our method, we conduct experiments on five real-world TSAD datasets and compare it with eight SotA methods. 

The results for the methods in AUC-ROC, AUC-PR, and best $F_1$ metrics are shown in Table \ref{tab:main_result}. Our method STEN is the best performer on all five datasets across the three metrics, indicating the importance and effectiveness of joint modeling of spatial-temporal normality. On average, STEN obtains an AUC-ROC of 0.997, an AUC-PR of 0.973, and an $F_1$ of 0.966. Specifically, in terms of the AUC-PR metric, STEN achieves an average improvement ranging from approximately 1.3\% to 31.6\% over eight SotA methods, which highlights the effectiveness of our method in the precision and recall rates of detecting anomalies. 
Impressively, our method marks a significant improvement on the SMAP dataset, achieving an improvement of 2\% to 34.7\% in the AUC-PR metric. 
Compared to other methods, AT exhibits relatively better detection performance on these datasets, utilizing advanced Transformer structures and concepts like association differences for effective normality modeling. Nevertheless, our method STEN still consistently outperforms AT, e.g., by a large margin on some challenging datasets like MSL and DSADS. 
This is mainly due to the additional spatial normality modeling in STEN, besides the temporal normality learning.

\begin{table}[!t]
  \centering
  \caption{Aff-$F_{1}$, $R_{\text{AUC-ROC}}$, $R_{\text{AUC-PR}}$, VUS-ROC, and VUS-PR results.}
   \scalebox{0.8}{%
    \begin{tabularx}{\textwidth}{
    ll
    >{\centering\arraybackslash}X
    >{\centering\arraybackslash}X
    >{\centering\arraybackslash}X
    >{\centering\arraybackslash}X
    >{\centering\arraybackslash}X}
    \toprule
    \multicolumn{1}{l}{\textbf{Dataset  }} & \textbf{Method} & \textbf{Aff-$F_{1}$} & \textbf{$R_{\text{AUC-ROC}}$} & \textbf{$R_{\text{AUC-PR}}$} & \textbf{VUS-ROC} & \textbf{VUS-PR} \\
    \midrule
    \multirow{4}[2]{*}{PSM} & NCA & 0.394  & 0.862  & 0.826  & 0.868  & 0.828  \\
          & DC & 0.583  & 0.817  & 0.839  & 0.805  & 0.826  \\
          & AT    & 0.507  & 0.952  & \textbf{0.959 } & 0.883  & 0.905  \\
          & \textbf{Ours} & \textbf{0.636 } & \textbf{0.973 } & 0.957  & \textbf{0.966 } & \textbf{0.945 } \\
    \midrule
    \multirow{4}[2]{*}{MSL} & NCA  & 0.586  & 0.947  & 0.755  & 0.949  & 0.760  \\
          & DC & 0.641  & 0.844  & 0.791  & 0.841  & 0.786  \\
          & AT    & 0.659  & 0.915  & 0.844  & 0.917  & 0.804  \\
          & \textbf{Ours} & \textbf{0.687 } & \textbf{0.955 } & \textbf{0.853 } & \textbf{0.957 } & \textbf{0.858 } \\
    \midrule
    \multirow{4}[2]{*}{SMAP} & NCA  & 0.662  & 0.958  & 0.838  & 0.957  & 0.837  \\
          & DC & 0.662  & 0.944  & 0.914  & 0.939  & 0.908  \\
          & AT    & 0.675  & 0.968  & 0.938  & 0.958  & 0.930  \\
          & \textbf{Ours} & \textbf{0.677 } & \textbf{0.985 } & \textbf{0.954 } & \textbf{0.985 } & \textbf{0.952 } \\
    \midrule
    \multirow{4}[2]{*}{Epilepsy} & NCA  & 0.585  & \textbf{0.965 } & 0.917  & 0.965  & 0.917  \\
          & DC & 0.655  & 0.770  & 0.808  & 0.773  & 0.811  \\
          & AT    & 0.691  & 0.935  & 0.936  & 0.939  & 0.940  \\
          & \textbf{Ours} & \textbf{0.746 } & 0.964  & \textbf{0.943 } & \textbf{0.966 } & \textbf{0.947 } \\
    \midrule
    \multirow{4}[2]{*}{DSADS} & NCA  & 0.650  & 0.835  & 0.779  & 0.839  & 0.785  \\
          & DC & 0.617  & \textbf{0.942 } & 0.829  & 0.943  & 0.834  \\
          & AT    & \textbf{0.682 } & 0.866  & \textbf{0.860 } & 0.870  & 0.864  \\
          & \textbf{Ours} & 0.657  & 0.940  & 0.856  & \textbf{0.944 } & \textbf{0.865 } \\
    \bottomrule
    \end{tabularx}%
    }%
  \label{tab:multi-result}%
\end{table}%

It is important to note that recent studies have had vigorous discussions on how to fairly evaluate the performance of TSAD methods. Although the results in Table \ref{tab:main_result} have already demonstrated that our method outperforms the SotA methods under the three traditional metrics, to have a more comprehensive and fair evaluation, we also use several recently proposed evaluation metrics. 
The evaluation results of our model under the new metrics are shown in Table \ref{tab:multi-result}, with the best competing models NCA, DC, and AT as our baselines. 
It is evident that on the PSM, MSL and SMAP datasets, our method outperforms all others across almost all new metrics. For the remaining two datasets, we still achieve superior results, with a minimal margin of less than 0.005 in all cases except Aff-$F_{1}$ on DSADS. These results re-affirm the improvement of our method over SotA methods.

\subsection{Ablation Study}
\noindent\textbf{Significance of the OTN and DSN Modules.} We then investigate the importance of the two components DSN and OTN. We remove each of them from our method STEN individually, resulting in two ablation variants: one with only DSN and another with only OTN. The results are presented in Table \ref{tab:ablation}, which demonstrate that both of the DSN and OTN modules have some major contributions to the model's superior detection performance. This is particularly true for the OTN module, which significantly increases the model's performance in terms of the $F_1$ score by an average of approximately 16.9\% over the five datasets. 

\vspace{0.3cm}
\noindent\textbf{OTN vs Error Prediction-based Approach in Modeling Temporal Normality.} Furthermore, to illustrate the effectiveness of OTN in capturing temporal patterns, we derive a variant of STEN that combines the DSN with a popular temporal pattern modeling module based on error prediction (EP) of the current timestamp. The experiment results reveal that the OTN module achieves improvements of approximately 7.3\% in AUC-PR and 10.1\% in the $F_1$ score, respectively. These results demonstrate that our OTN module is significantly more effective compared to the popular error prediction methods, which is mainly due to the fact that the OTN module allows the modeling of normal patterns in multiple diverse temporal contexts, contrasting to the single temporal context in the existing error prediction method.

\begin{table}[!t]
  \centering
  \caption{Ablation study results of STEN. EP represents the popular error prediction-based approach for temporal pattern modeling. }
   \scalebox{0.8}{%
    \begin{tabular}{ccccccccccccc}
    \toprule
          &   & \multicolumn{2}{c}{\textbf{PSM}} & \multicolumn{2}{c}{\textbf{MSL}} & \multicolumn{2}{c}{\textbf{SMAP}} & \multicolumn{2}{c}{\textbf{ Epilepsy }} & \multicolumn{2}{c}{\textbf{ DSADS }} \\
              \cmidrule(lr){3-4} \cmidrule(lr){5-6} \cmidrule(lr){7-8} \cmidrule(lr){9-10} \cmidrule(lr){11-12}
   \textbf{DSN} & \textbf{OTN} & \textbf{AUC-PR} & \textbf{$F_1$} & \textbf{AUC-PR} & \textbf{$F_1$} & \textbf{AUC-PR} & \textbf{$F_1$} & \textbf{AUC-PR} & \textbf{$F_1$} & \textbf{AUC-PR} & \textbf{$F_1$} \\
    \midrule
    \Checkmark & \XSolidBrush & 0.913  & 0.902  & 0.763  & 0.752  & 0.847  & 0.794  & 0.794  & 0.808  & 0.790  & 0.728  \\
    \XSolidBrush & \Checkmark & 0.995  & 0.975  & \textbf{0.946 } & 0.919  & 0.990  & 0.963  & 0.979  & 0.978  & 0.929  & 0.906  \\
    \Checkmark & \Checkmark & \textbf{0.995}  & \textbf{0.986 } & 0.941  & \textbf{0.944 } & \textbf{0.991 } & \textbf{0.974 } & \textbf{0.991 } & \textbf{0.991 } & \textbf{0.948 } & \textbf{0.934 } \\\hline
    \multicolumn{2}{c}{DSN + EP}& 0.984  & 0.952  & 0.930  & 0.897  & 0.921  & 0.856  & 0.957  & 0.939  & 0.707  & 0.681  \\
    \bottomrule
    \end{tabular}%
    }%
  \label{tab:ablation}%
\end{table}%

\subsection{Qualitative Analysis }

\begin{figure}[!t]
\includegraphics[width=\textwidth]{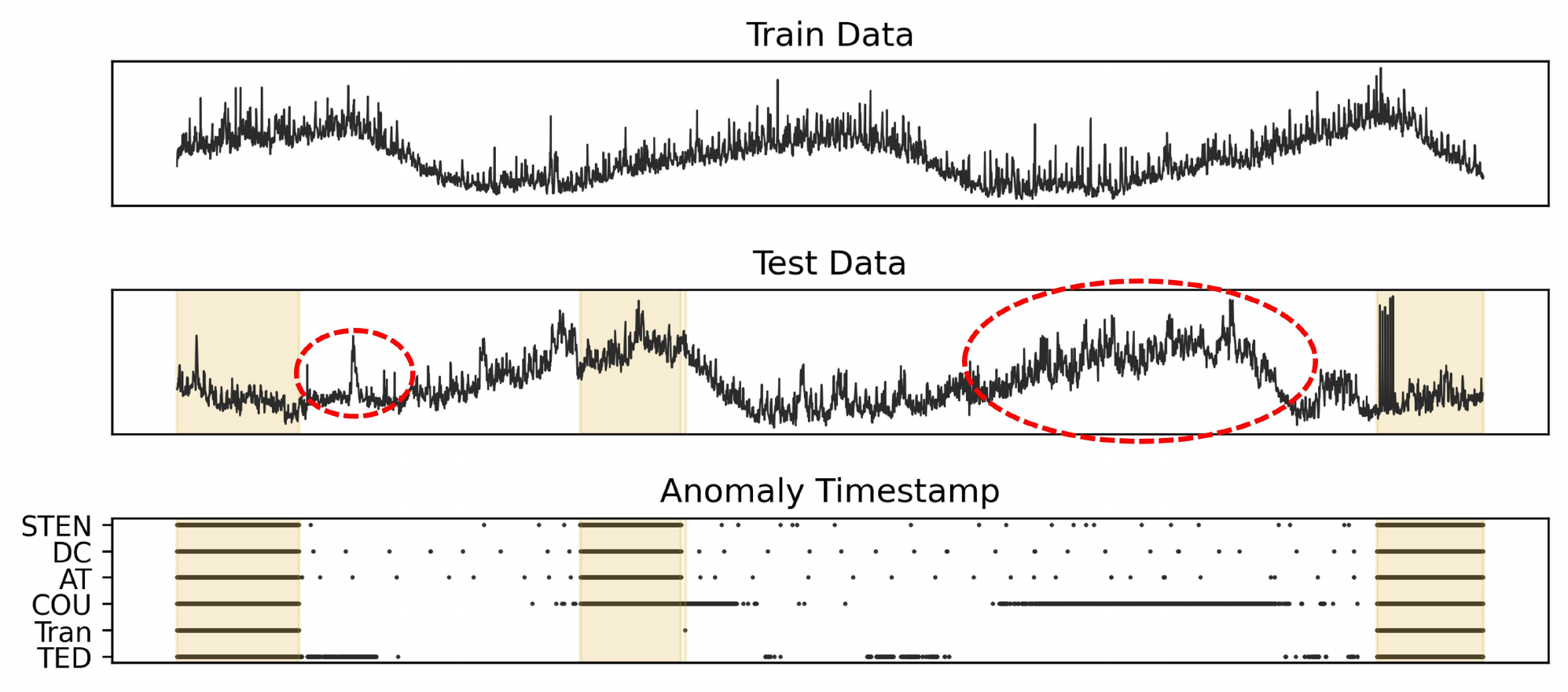}
\caption{\textbf{(Top to Bottom)} Training/testing segments of the PSM dataset, and the anomalous timestamps (marked by small dots) predicted by STEN and the five best-performing competing methods. The ground truth anomalous segments are highlighted in gold. The most prominent false-positive data segments are encircled in red dashed lines.} 
\label{fig:visualization}
\end{figure}

We visualize the results to further showcase the anomaly detection capabilities of STEN. Fig. \ref{fig:visualization} presents the data from the PSM dataset (feature id 22), along with the anomaly timestamps detected by STEN and its competing methods. This dataset exhibits complex temporal patterns, which may lead to false positives by many algorithms during detection. Compared to the competing methods, our method STEN can detect all three anomalies while at the same time having the least false positives. 
For example, COU and TED incorrectly label test data segments that are similar to the normal patterns in the training data as anomalies. 
Meanwhile, both Tran and TED miss an anomaly, i.e., the second anomalous segment. 
DC and AT illustrate comparable performance in detecting all three anomalies, but they still produce more false positives than our method. This advatange is due to the spatial normality modeling of STEN, which is often ignored in existing methods like DC and AT.

\subsection{Sensitivity Analysis}

\begin{figure}[!t]
  \centering
  \begin{subfigure}[b]{0.45\textwidth}
    \includegraphics[width=\textwidth]{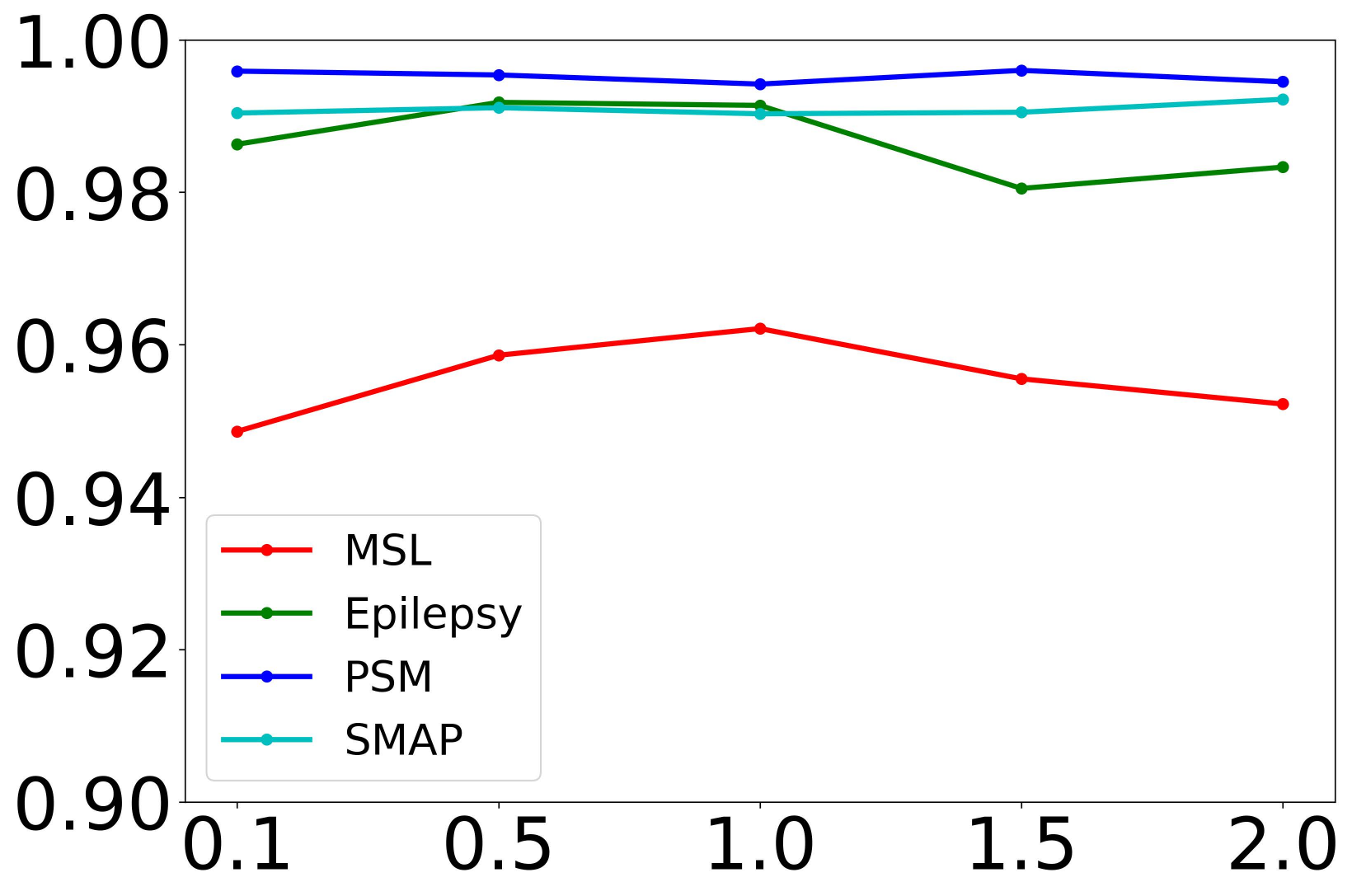}
    \caption{$\alpha $}
    \label{fig:sen_sub1}
  \end{subfigure}
  \begin{subfigure}[b]{0.45\textwidth}
    \includegraphics[width=\textwidth]{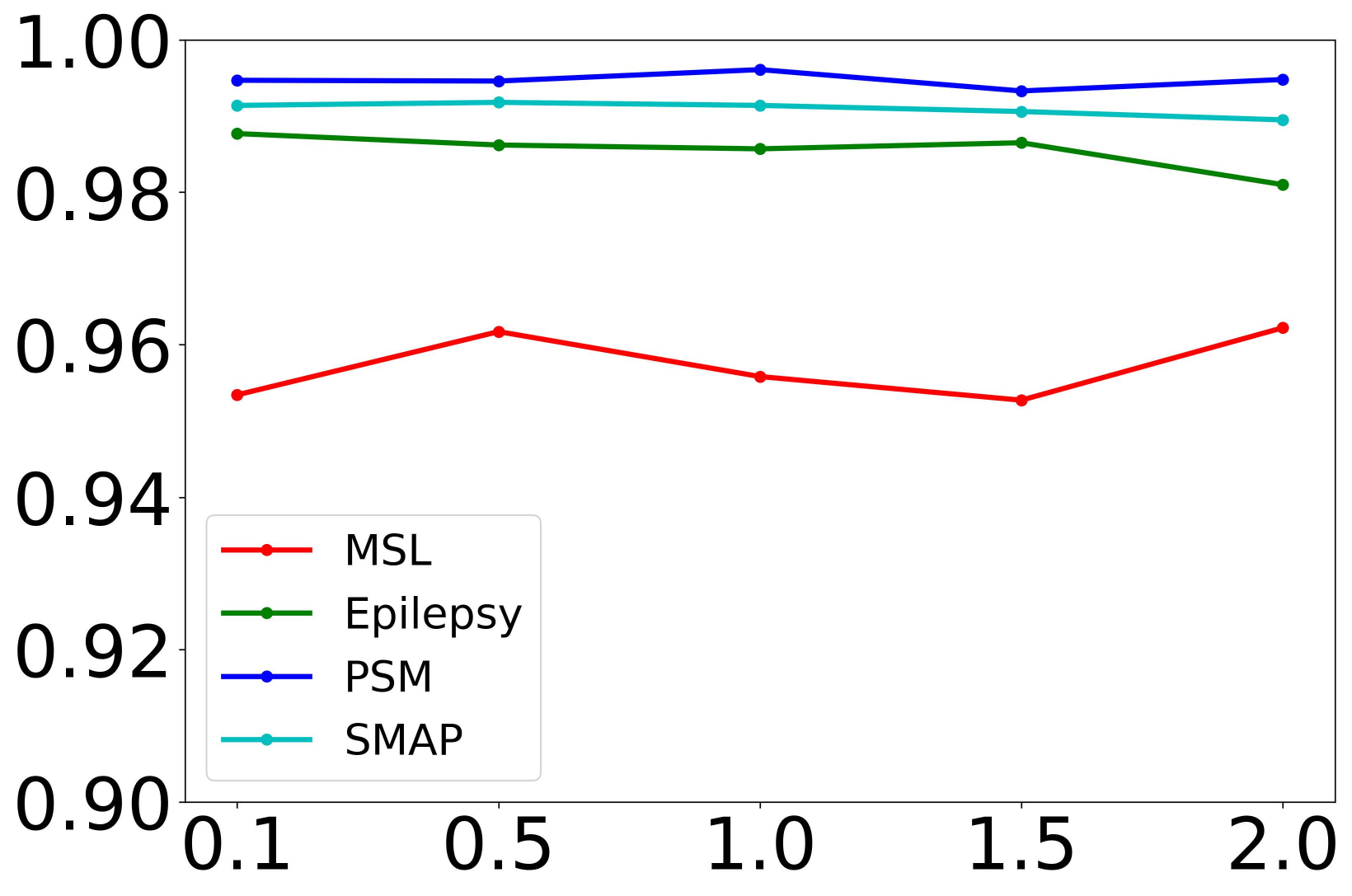}
    \caption{$\beta$}
    \label{fig:sen_sub2}
  \end{subfigure}
  \begin{subfigure}[b]{0.45\textwidth}
    \includegraphics[width=\textwidth]{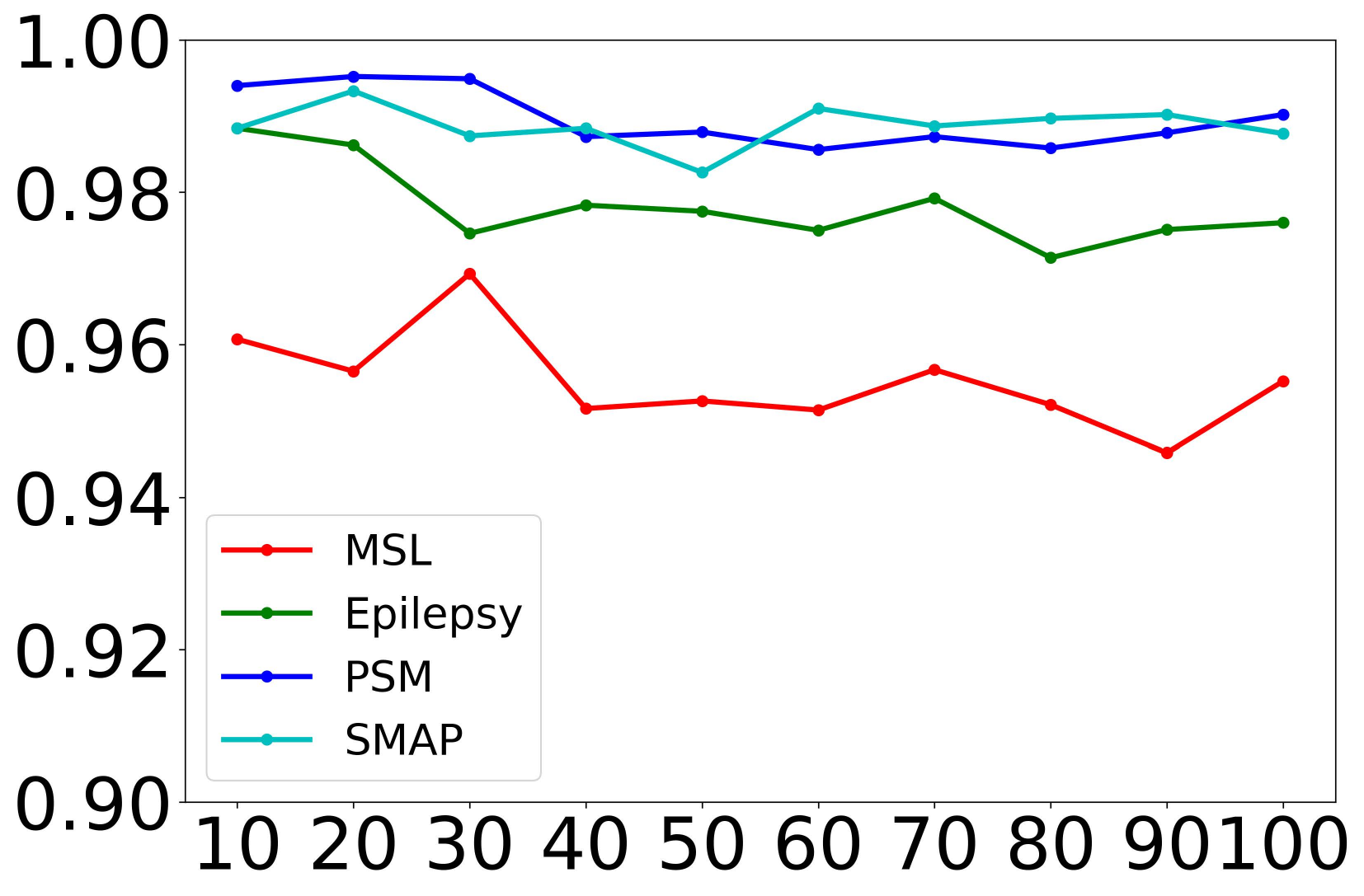}
    \caption{$l$}
    \label{fig:sen_sub3}
  \end{subfigure}
  \begin{subfigure}[b]{0.45\textwidth}
    \includegraphics[width=\textwidth]{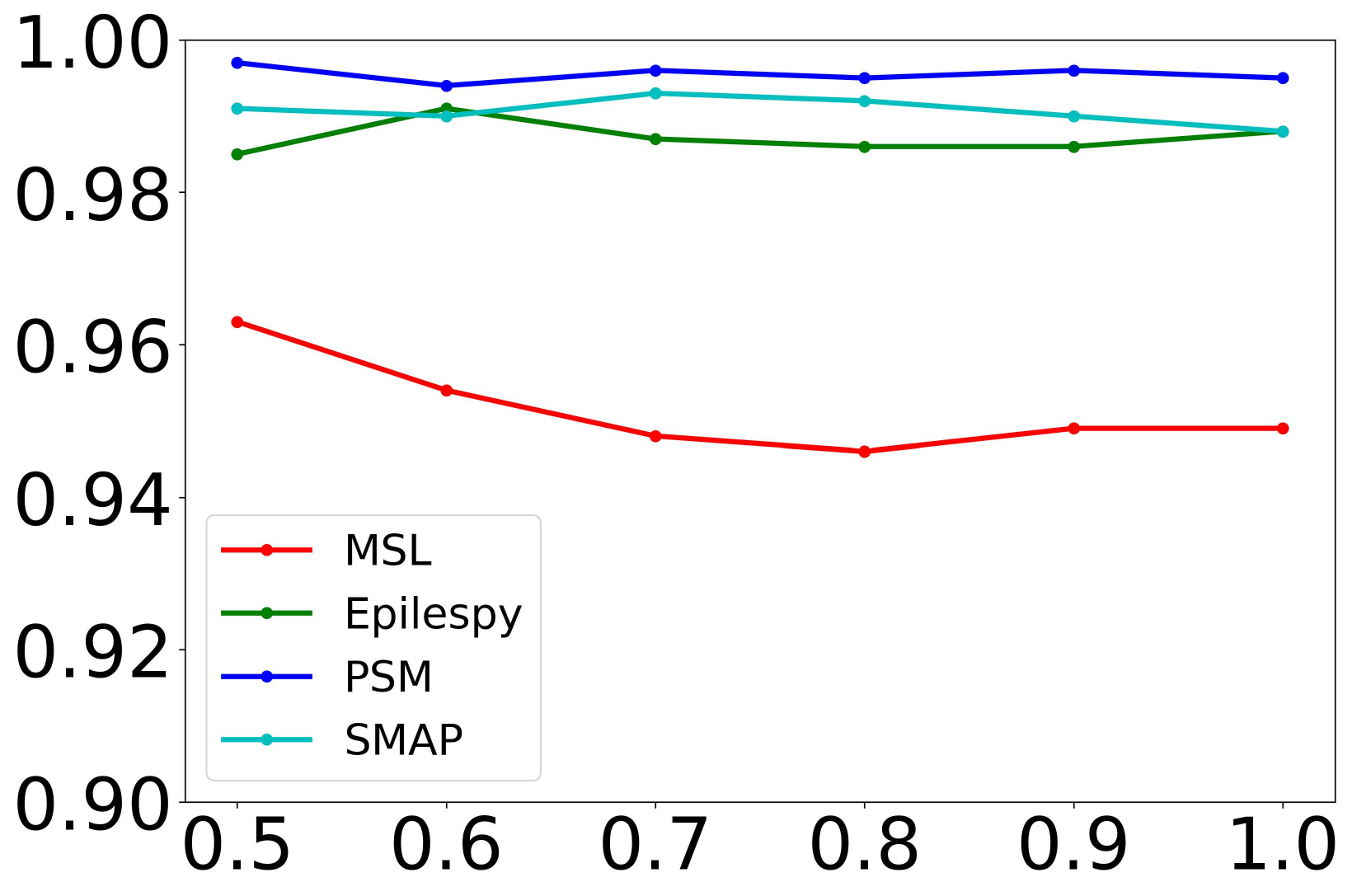}
    \caption{$\delta $}
    \label{fig:sen_sub4}
  \end{subfigure}
  \caption{AUC-PR results of STEN w.r.t different hyperparameters.}
  \label{fig:Sensitivity}
\end{figure}

We also conduct sensitivity analysis on four key hyperparameters. The $F_{1}$ score and AUC-PR results have a similar trend and we report the AUC-PR results in Fig. \ref{fig:Sensitivity} due to page limitation. Fig. \ref{fig:Sensitivity}(a) illustrates the sensitivity of the hyperparameter regulating the loss $\mathcal{L}_{dsn}$, i.e., $\alpha $, in the model’s overall loss function. This analysis aims to investigate the model’s capability to capture the normal patterns of data when applying different weights to our two modules. The results indicate that the model’s performance remains stable across multiple benchmark datasets within a relatively large value range. Fig. \ref{fig:Sensitivity}(b) focuses on the weight $\beta$ of ${Score}_{{dsn}}(R_{i})$, examining the contribution of the two modules to the anomaly scoring. The results suggest that varying $\beta$ does not significantly affect the model’s detection performance, showcasing the model’s robustness. 
We also experiment with different sub-sequence lengths $l$. The results are shown in Fig. \ref{fig:Sensitivity}(c), from which it is clear that our method achieves stable detection performance with sub-sequences of various lengths ranging from 10 to 100.
Furthermore, the anomaly threshold $\delta $ serves as a hyperparameter that distinguishes anomalies from normal fluctuations in the data. 
Fig. \ref{fig:Sensitivity}(d) illustrates the experimental results when $\delta $ varies between 0.5 and 1. It can be observed that our method performs stably w.r.t. changes in $\delta$.

\subsection{Time Efficiency}

We compare training time of neural network-based models across various datasets to investigate their time efficiency. 
Due to the obviously high algorithmic complexity of GDN and MSCRED, we select six methods with comparable training duration for comparison, as shown in Table \ref{tab:efficiency}. Unlike previous methods, our model employs two modules (DSN and OTN) to train and combine sub-sequences of different length, potentially increasing the computational complexity. 
Although not showing the most efficient performance, our method illustrates significantly better detection accuracy than the competing methods.

\begin{table}[htbp]
  \centering
  \caption{Run times (in seconds) in multiple datasets. The best results are indicated in bold, while the worst results are underlined.}
  \scalebox{0.9}{%
    \begin{tabular}{lccccc}
    \toprule
    \textbf{Methods} & \makebox[0.15\textwidth][c]{\textbf{PSM}} & \makebox[0.15\textwidth][c]{\textbf{MSL}} & \makebox[0.15\textwidth][c]{\textbf{SMAP}} & \makebox[0.15\textwidth][c]{\textbf{Epilepsy}} & \makebox[0.15\textwidth][c]{\textbf{DSADS}} \\
    \midrule
    TED   & 308.2 & 140.0   & 317.3 & 76.0    & \underline{1646.3} \\
    Tran  & 144.5 & 70.3  & 152.8 & 35.2  & 755.1 \\
    NCA   & \underline{711.3} & 330.3 & 746.8 & \underline{179.0}   & 421.5 \\
    COU   & 282.5 & 127.0   & 286.6 & 71.5  & 1440.2 \\
    DC    & 508.6 & \underline{380.8} & \underline{967.1} & 98.3  & 1536.5 \\
    AT    & \textbf{11.7} & \textbf{11.1} & \textbf{12.6} & \textbf{10.8} & \textbf{15.4} \\
    \midrule
    \textbf{Ours}  & 214.9 & 137.3 & 477.8 & 44.4  & 483.3 \\
    \bottomrule
    \end{tabular}%
    }
  \label{tab:efficiency}%
\end{table}%

\section{Conclusion}
This article introduces a novel TSAD method named STEN. Unlike existing methods that solely focus on the temporal normality of time series data, STEN incorporates the spatial normality of the data as well, enabling a more comprehensive normality learning of the time series data. This enables STEN to achieve greater discriminative feature representations in distinguishing abnormal sequences from the normal ones. 
In STEN, we design two pretext tasks to extract spatial-temporal features of time series data. The order prediction-based temporal normality learning (OTN) models the temporal dependencies of the data by modeling the distribution of sub-sequence order, while the distance prediction-based spatial normality learning (DSN) learns the spatial relation of sequence pairs in the projected space in the form of distance prediction. 
Extensive experiments demonstrate that STEN exhibits superior TSAD performance, outperforming eight SotA algorithms on five benchmark datasets, with the effectiveness of each of the proposed two modules justified in our ablation study.

\begin{credits}
\subsubsection{\ackname} This work was supported in part by the Key Cooperation Project of the Chongqing Municipal Education Commission under Grants HZ2021017 and HZ2021008, in part by the National Natural Science Foundation of China under Grant 62102442.

\subsubsection{\discintname}
The authors have no competing interests to declare that are relevant to the content of this article.
\end{credits}

\end{document}